\documentclass[runningheads]{llncs}

\usepackage{accv}

\usepackage{accvabbrv}
\usepackage{multirow}
\usepackage{graphicx}
\usepackage{booktabs}

\usepackage[accsupp]{axessibility}  

\usepackage[pagebackref,breaklinks,colorlinks,citecolor=accvblue]{hyperref}

\usepackage{orcidlink}

\begin{document}

\title{Telling Stories for Common Sense Zero-Shot Action Recognition} 


\author{Shreyank N Gowda\inst{1} \and
Laura Sevilla-Lara\inst{2}}

\authorrunning{S. N. Gowda et al.}

\institute{University of Nottingham, UK \and
University of Edinburgh, UK\\
\email{Shreyank.Narayanagowda@nottingham.ac.uk}}

\maketitle

\begin{abstract}
Video understanding has long suffered from reliance on large labeled datasets, motivating research into zero-shot learning. Recent progress in language modeling presents opportunities to advance zero-shot video analysis, but constructing an effective semantic space relating action classes remains challenging. We address this by introducing a novel dataset, {\em Stories}, which contains rich textual descriptions for diverse action classes extracted from WikiHow articles. For each class, we extract multi-sentence narratives detailing the necessary steps, scenes, objects, and verbs that characterize the action. This contextual data enables modeling of nuanced relationships between actions, paving the way for zero-shot transfer. We also propose an approach that harnesses Stories to improve feature generation for training zero-shot classification. 
Without any target dataset fine-tuning, our method achieves new state-of-the-art on multiple benchmarks, improving top-1 accuracy by up to 6.1\%. We believe Stories provides a valuable resource that can catalyze progress in zero-shot action recognition. The textual narratives forge connections between seen and unseen classes, overcoming the bottleneck of labeled data that has long impeded advancements in this exciting domain. The data can be found here: \url{https://github.com/kini5gowda/Stories}.
\end{abstract}

\section{Introduction}
\label{sec:intro}

Action recognition technology has improved remarkably over the years, with methods becoming more accurate. However, one of the main challenges that remain today is the dependency of these methods on annotated data for novel categories. The availability of large labeled datasets like ImageNet \cite{imagenet} has played a pivotal role in propelling the field of supervised learning seeing better than human level performance on image classification \cite{resnet,vit,colornet} tasks. In practice, obtaining large amounts of annotated examples for each new class we aim to recognize is not realistic and not practical~\cite{watt}. This is particularly true as we grow the number of classes and wish to incorporate more flexible, natural language, for example, for retrieval. This general problem has led to research in the zero-shot (ZS) domain. 

\begin{figure}[t]
    \centering    \includegraphics[width=0.85\linewidth]{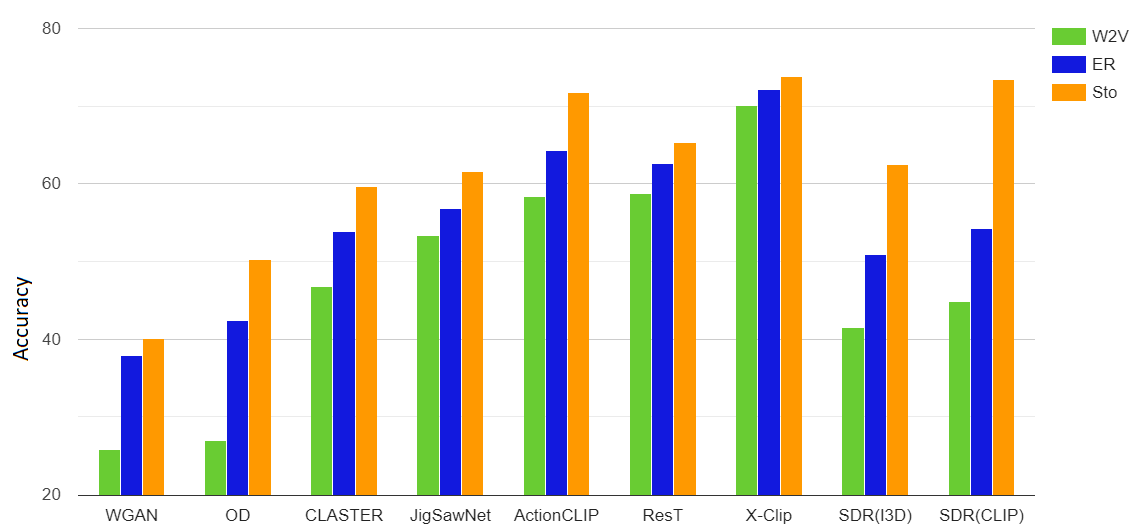}
\caption{Comparison of accuracy across state-of-the-art ZS approaches using different semantic embeddings: the proposed {\em Stories}, word2vec (W2V) and elaborative definitions (ER), on UCF101. Using the proposed {\em Stories~} to create the semantic space of class labels improves the performance by a large margin across all methods, showing that it is model-agnostic. }
    \label{fig:teaser}
\end{figure}

In the typical ZS setting, there are seen classes that contain visual examples and their class label, and there are unseen classes where only the class labels are available at training time. Given a visual sample of the unseen set at test time, the task is to output the corresponding class label. Approaches typically learn a mapping from the visual space to the class labels using the seen classes and leverage that mapping in different ways to approximate the mapping in the unseen classes. Some examples of solutions include learning to map visual information and class labels to the same space or learning to generate visual features using a generative adversarial network (GAN)~\cite{clswgan}.

One of the underlying assumptions of these approaches is that the distances between the data points are meaningful both in the visual space and the semantic space. In other words, data points that are close together should be similar in content across both seen and unseen classes. In visual space, this tends to happen naturally as related classes will share objects, scenes, etc. For example, if we compare classes such as ``penalty shot" and ``playing soccer", they will share the ball, the soccer field, etc. However, in the space of action labels, which we also refer to as semantic space, this property is not straightforward to achieve. While some similar classes will contain overlapping terms (such as ``horse-back riding" and ``horse racing"), some others might be similar but not contain overlapping words (such as the previous example of ``penalty shot"  and ``playing soccer"). This makes the step of transferring knowledge between seen and unseen classes harder. Previous efforts to improve the semantic space of class labels have included the use of manually annotated attributes or embedding functions trained on language corpora, such as word2vec~\cite{word2vec}, sentence2vec~\cite{sentence2vec} and using definitions of actions~\cite{er}. 

In this work, we address the problem of building a meaningful space of action labels by leveraging the story around each action. In particular, we use the descriptions of the steps needed to achieve each action, which are the {\em Stories} around this action and encode them using a language embedding. These steps typically contain the objects, tools, scenes, verbs, adjectives, etc., associated with the action label. One could think of all these additional pieces of information around an action as the ``common sense" of associations humans would typically consider. For example, in the case of the penalty shot, the steps would describe to first place the ball, run the hand through the grass, fluff the ball, take some steps back, kick, etc. When playing soccer, the steps include kicking the ball with the inside of your shoe for short passes across the grass, tapping the ball from foot to foot, etc. When we compare the stories of steps around these two classes, the overlap of terms becomes much more obvious. It is more likely that these related classes are closer in semantic space, facilitating the transfer of knowledge between seen and unseen classes. We show that this relatively simple approach to creating a semantic space is extremely effective across datasets and methods, improving performance by up to 20\% compared to the standard word2vec~\cite{word2vec}. Figure~\ref{fig:teaser} shows that all state-of-the-art methods improve significantly and therefore the proposed semantic embedding is general and model-agnostic.

Finally, we leverage these Stories to go beyond simply learning semantic representations, to actually generating additional features that improve the semantic space further.  
%
We follow a feature-generating approach~\cite{OD,clswgan} using a GAN~\cite{goodfellow2020generative} to synthesize visual data points from these semantic embeddings. These synthetic visual data points are then used to learn a mapping from visual to semantic space in the unseen classes.  
We show that this method improves state-of-the-art by an additional 6.1\%. We observe a strong trend of feature generating networks benefiting particularly from using Stories. The largest jumps using Stories are in feature generating methods (SDR-I3D, SDR-CLIP and OD), as shown in Figure~\ref{fig:teaser}. 


\section{Related Work}
\label{sec:related_work}

\paragraph{Fully Supervised Action Recognition.}

In this setting, there is a large amount of training samples with their associated labels and the label spaces are the same at train and test time. Much of the advances in this area of research is often used on ZS. Early work in deep learning for action recognition used many tools to represent the spatio-temporal nature of videos, including 3D CNNs~\cite{i3d}, 2D~CNNs with temporal modules \cite{tsm,kim2022capturing,gowda2017human}, 2D~CNNS with relational modules \cite{smart} and two-stream networks~\cite{simonyan2014two}. More recently, the transformer architecture \cite{vit} is particularly well suited to represent sequential information and has been successfully adapted to the video domain~\cite{timesformer,vivit,videoswin,mvit}. 
While these are extremely powerful tools, they are difficult to train with limited data. Instead, we use the standard I3D \cite{i3d} as our backbone feature generator to compare directly with recent state-of-the-art papers \cite{claster,OD}.

\paragraph{Zero-Shot Action Recognition.}

Early work~\cite{rohrbach12eccv} in 
this setting used script data in cooking activities to transfer to unseen classes. Considering each action class as a domain, Gan et al. \cite{gan2016learning} address the identification of semantic representations as a multisource domain generalization problem. To obtain semantic embeddings of class names, label embeddings such as word2vec~\cite{word2vec} has proven popular as only class names are needed. Some approaches use a common embedding space between
video features and class labels~\cite{xu2016multi,xu2017transductive},
error-correcting codes \cite{qin2017zero}, pairwise relationships between classes \cite{gan2016concepts}, interclass relationships \cite{gan2015exploring}, out-of-distribution detectors~\cite{OD}, and graph neural networks~\cite{gao2019know}. 
Recently, it was seen that clustering of joint visual-semantic features helped obtain better representations~\cite{claster} for ZS action recognition.  Similar to CLASTER, ReST \cite{rest} jointly encodes video data and text labels for ZS action recognition. In ResT, transformers are used to perform modality-specific attention. JigSawNet \cite{jigsaw} also models visual and textual features jointly but decomposes videos into atomic actions in an unsupervised manner and bridges group-to-group relationships between visual and semantic representations instead of the one-to-one relationships that CLASTER and ReST do. Unlike JigSawNet, that works on the visual features, we create enriched textual descriptions by decomposing actions into a series of steps and hence obtain richer semantic features. Other solutions to deal with limited labeled data include augmentation~\cite{L2A}, relational modules~\cite{trx}.

\begin{figure*}[t]
    \centering
    \includegraphics[width=0.9\linewidth]{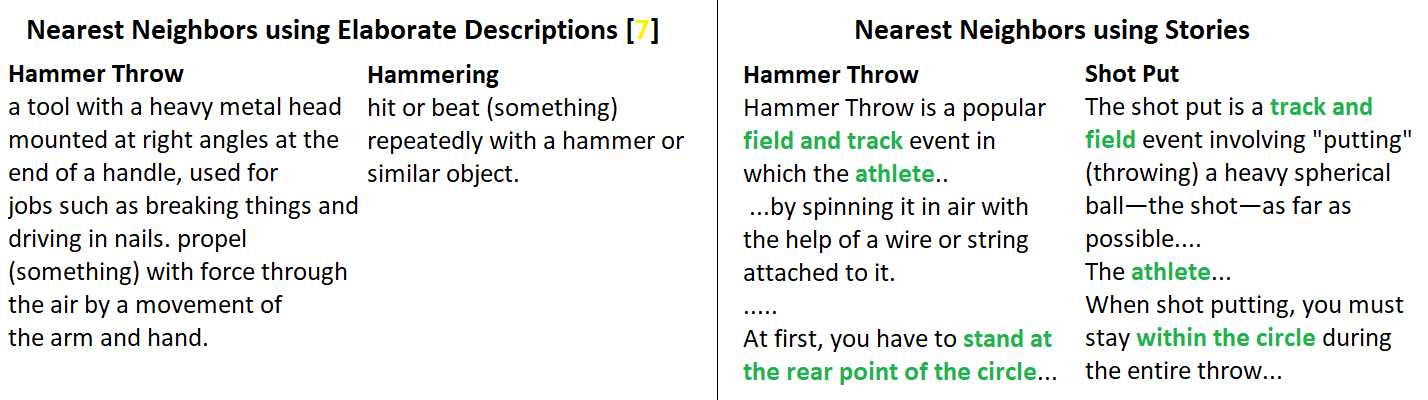}
    \caption{Comparing nearest neighbors using Stories. We see an example where ER fails and Stories provides more context and helps in obtaining better neighbors. This is one example of where ER fails, there are multiple such examples. Dataset is UCF101.}
    \label{fig:er_comparison}
\end{figure*}

\paragraph{Semantic Embeddings.}

To obtain semantic embeddings of action class labels, earlier works use word2vec \cite{word2vec} directly on the class labels. However, word2vec averages the embedding for class labels with multiple words, giving equal weight to each word. This causes class names to lose context. For example, the class ``pommel horse" is a gymnastic class which does not involve the animal horse. Unfortunately, using word2vec makes the embedding of that class close to ``horse riding" or ``horse racing" in the word2vec space, even though they are not gymnastics sports. A recent solution is to use elaborate descriptions~\cite{er} based on the principle of Elaborative Rehearsals (ER), which replace each class name with a class definition. An object description was also used to describe the objects in the particular action. This resulted in a significant boost in performance. Still, ER uses descriptions of each word in a class label independently to create the semantic space, which potentially leads to errors. 

\paragraph{Feature Generating Networks for ZS.}

Bucher et al.~\cite{bucher2017generating} proposed to bridge the gap between seen and unseen classes by synthesizing features for the unseen classes using four different generator models. Xian et al.~\cite{clswgan} trained their generators by using a conditional generative adversarial network (GAN)~\cite{mirza2014conditional}. In contrast, Verma et al.~\cite{verma2018generalized} trained a variational autoencoder. Mishra et al.~\cite{GGM2018} introduced the generative approach to the domain of ZS action recognition. Their approach models each action class as a probability distribution in the visual space. Mandal et al.~\cite{OD} modified the work by Xian et al.~\cite{clswgan} to work directly on video features. They also introduced an out-of-distribution detector to help with the generalized zero-shot learning setting. Here we propose a variant of the work on out-of-distribution detectors, which suffers less from long converging times yet it improves its accuracy. More recently Gowda et al.~\cite{gowda2023synthetic}, proposed to have a selector that selects generated features based on their importance on training the classifier rather than on realness of the features.

\section{The Stories Dataset}

Research in semantic representation of action labels has shown over the years that more sophisticated representations help to build a meaningful semantic space for ZS action recognition. Here we go beyond previous work by representing not only the class label but the story around it. This is, we build a representation that captures all the steps needed to perform the action, which include objects, verbs, etc, typically associated with that action. We call these representations {\em Stories~}, and we now describe how we build them. 

\subsection{Building the Stories Dataset}
\begin{figure}[t]
\centering
\includegraphics[width=0.48\textwidth]{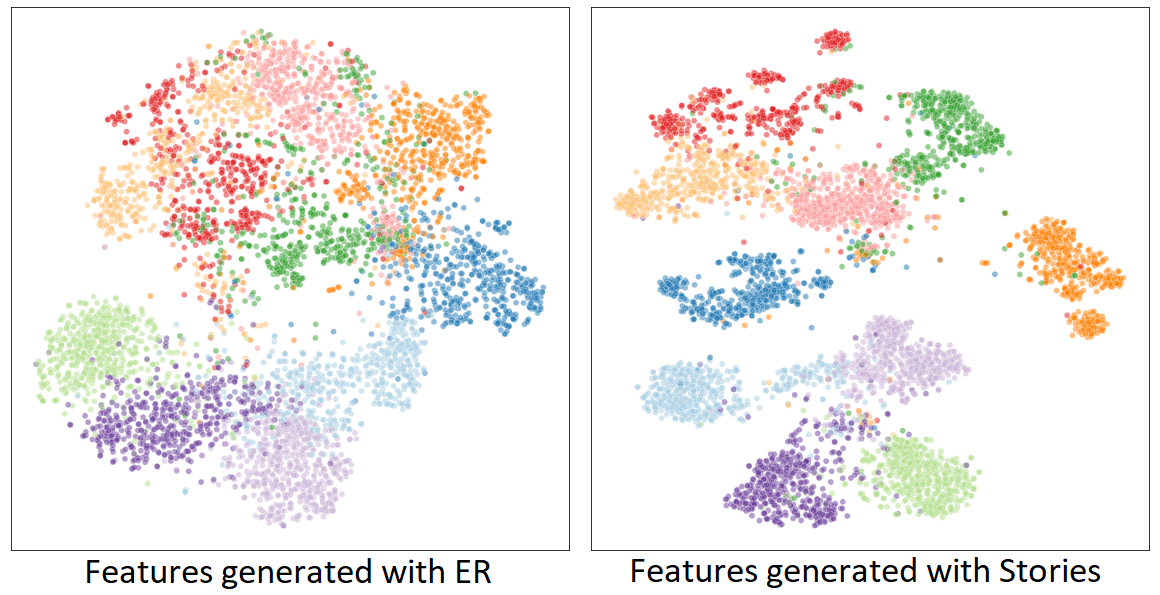}
\caption{Visualization of the features generated from the \emph{embedding} vs ER, using t-SNE. We observe that the samples of each class instance, depicted in a single color, are better clustered together, pointing to a more semantically meaningful space.}
\label{fig:tsne}
\end{figure}

We leverage textual descriptions of actions from \textit{WikiHow}\footnote{\url{https://www.wikihow.com/}}, a website that gives instructions on how to perform actions. These instructions consist of long paragraphs that describe each step in completing the action. For example, for the action classes in the HMDB51 dataset, the \textit{WikiHow} articles contain an average of 9.8 steps, ranging from 4 to 20 steps. The most closely related work to us, ER~\cite{er} uses a single or at max two sentences per class. Instead, for the proposed Stories~, the average number of lines is 14.4 for the classes in the Olympics dataset, 9.6 for HMDB51, 13.2 for UCF101, and 13.5 Kinetics. 
These rich descriptions inherently contain information of the objects needed to perform an action. For example, the class `biking' has a paragraph that explains where to bike, what equipment you need, and the steps to do it. In comparison, ED~\cite{er} has only a single line definition of the class.

Not all classes have articles on \textit{WikiHow}. For example, if we search for ``draw sword" (a class in UCF101), we will get instructions on how to paint or sketch swords instead of the steps needed to essentially remove a sword from its sheath. Hence, collecting clean, meaningful articles requires a more complex process than a simple search. After scraping the articles from \textit{WikiHow} corresponding to all classes, we use sentence-BERT encoders~\cite{para-bert} to represent the sentences in the article and use cosine similarity to find the 25 that are the most similar to the class definition~\cite{er}. 

Next, we manually check the sentences for each class. If we find a mismatch between the article and the action class, as was the case with the ``draw sword" example, we do a manual search to pick the most relevant article from other sources such as \textit{Wikipedia}. However, these alternative articles do not tend to contain the sequence of steps and hence need more manual intervention to order the sentences into a sequence of steps. 

We finally clean each story by re-arranging the sentences in sequential order and removing irrelevant sentences. In total, we had 6 people who manually cleaned the descriptions after the initial stage of noisy collection and a further 10 who verified the descriptions. This was done using the Prolific\footnote{\url{https://www.prolific.co/}} platform. The time taken for the cleaning Stories for UCF101, HMDB51 and Kinetics was 7.2 hours, 3.3 hours and 25.3 hours on average respectively.
We followed this process to create a dataset of these textual representations for classes in UCF101, HMDB51, Olympics, and Kinetics-400, as they are the most commonly used datasets for ZS action recognition.

\subsection{Learning from Stories}


In order to test the impact of using richer semantic representations, we used Stories~ as input to the state-of-the-art methods in ZS. To provide fair points of comparison, we also supply the standard word2vec~\cite{word2vec} embeddings as well as the more recently introduced ER~\cite{er} embeddings to these same models. By visually depicting the performance gains achieved by the models when using our proposed embeddings instead of the word2vec or ER embeddings in Fig.~\ref{fig:teaser}, we are able to clearly demonstrate the substantial improvements obtained through our approach. It increases up to 21\% compared to the widely used word2vec embeddings and 11.8\% over the ER embeddings. This comprehensive experimental analysis highlights how our semantically richer embeddings can notably boost performance across a diverse range of state-of-the-art models and is hence model agnostic.

We visualize the effect of using ER and Stories~ to generate features, using the t-SNE~\cite{tsne} for 10 classes in UCF101 (Fig.~\ref{fig:tsne}), all related to gymnastics and therefore easier to confuse. 
We see that using Stories~ helps keep a more meaningful neighborhood for visual instances, and keep classes apart. The visualization also shows why ER fails as there is not enough information to clearly distinguish classes.

\subsection{Why Are Stories Necessary?}
The proposed Stories~ dataset produces semantic embeddings for the class labels that are much more meaningful than previous work, and that reflects in the experimental improvement shown in Fig.~\ref{fig:teaser}. Here we delve deeper into what properties make Stories~ superior.

\paragraph{Capturing meaning of words jointly.}
Previous work~\cite{word2vec,er} represents a class label by computing a representation for each word in the class label and then computing the average. \footnote{In the case of ER, this is done for some of the classes.} While this in general is a sensible choice, it leads often to errors caused by words that have multiple meanings depending on the context. Fig.~\ref{fig:er_comparison} illustrates this issue with an example.  We use the class ``Hammer Throw" from the UCF101~\cite{ucf101} dataset, which is a sporting event in which the athlete throws a spherical object. If we retrieve the nearest neighbor with ER~\cite{er}, we obtain the class ``Hammering", which is not actually related in meaning but both contain the description of the tool hammer. However, if we use Stories~, the nearest neighbor is ``Shot Put", which is also a sporting event where the athlete throws a spherical object. Similar problems happen for example with classes such as ``Sword Exercise", or ``Swing Baseball". Overall, this shows the need for the more sophisticated joint description that Stories~ provides, instead of the per-word definition of ER.

\paragraph{Size of dataset.}

We also compare Stories to ER from a statistical perspective. We have briefly mentioned before that Stories~ contains much more detailed descriptions of the classes. Here we look at the numerical difference between ER and Stories~, shown in 
Table~\ref{tbl:stat}. The number of sentences is one order of magnitude larger, going from over 1 sentence on average per class, to over 10 sentences on overage, depending on the dataset. This ratio is also consistent in the number of nouns verbs, etc. 

\paragraph{Diversity of dataset.} Another aspect that increases the specificity of the class descriptions in addition to the size of the dataset is the diversity of the vocabulary. We look at the number of unique words in Table~\ref{tbl:stat} and observe that Stories~ contains more unique words than ER in all datasets and that the difference is particularly remarkable in smaller datasets. We argue that this diversity contributes to representing each class label in a more unique way, leading to a more sparse yet meaningful space.

\paragraph{Cleaning data manually.}
Generally, training with more data tends to produce better results. There often is a tension between using a smaller amount of clean data or a larger amount of noisy data. Here we have explored the effect of cleaning the data of Stories~ manually. 
We see that using the noisy version of the dataset (see Supplementary for more details) improves the performance over ER across methods but is still consistently worse than the cleaned version, even though it is roughly twice as large. This shows that the effect of cleaning up the data manually is not trivial. 
In the ER work, however, there exist errors that can be solved through manual revision. For example, ``Table Tennis Shot" has the definition ``put (food) into the mouth and chew and swallow it" which clearly corresponds to the wrong action class. 

\begin{table}[t]
\centering
\begin{tabular}{|c|c|c|c|c|c|c|c|}
\hline
Method & D & N & V & A & Ad & UW & S \\
\hline \hline
Stories & K & 68.5 & 37.9 & 7.1 & 0.04 & 35.1 & 13.5 \\
ER & K & 9.2 & 3.8 & 1.0 & 0.01 & 30.2 & 1.8 \\
\hline
Stories & U & 69.6 & 37.2 & 7.7 & 0.1 & 32.1 & 13.2 \\
ER & U & 9.7 & 4.3 & 1.4 & 0.06 & 29.5 & 1.8 \\
\hline
Stories & H & 59.2 & 33.4 & 6.5 & 0.3 & 34.5 & 9.6 \\
ER & H & 5.8 & 2.1 & 0.9 & 0.05 & 17.7 & 1.2 \\
\hline
Stories & O & 68.9 & 38.6 & 11.8 & 2.0 & 57.2 & 14.4 \\
ER & O & 9.1 & 4.2 & 1.2 & 0.3 & 31.4 & 1.7 \\
\hline
\end{tabular}
\caption{Statistical comparison of Stories to ER. We observe a larger number of Nouns (N), Verbs (V), Adverbs (A), Adjectives (Ad), Unique Words (UW), and Sentences (S) are averages across all the classes in a particular dataset (D). 'K' refers to Kinetics-600, 'U' to UCF101, 'H' to HMDB51, and 'O' to Olympics.}
\label{tbl:stat}
\end{table}

\begin{figure*}
        \centering
        \includegraphics[width=0.8\linewidth]{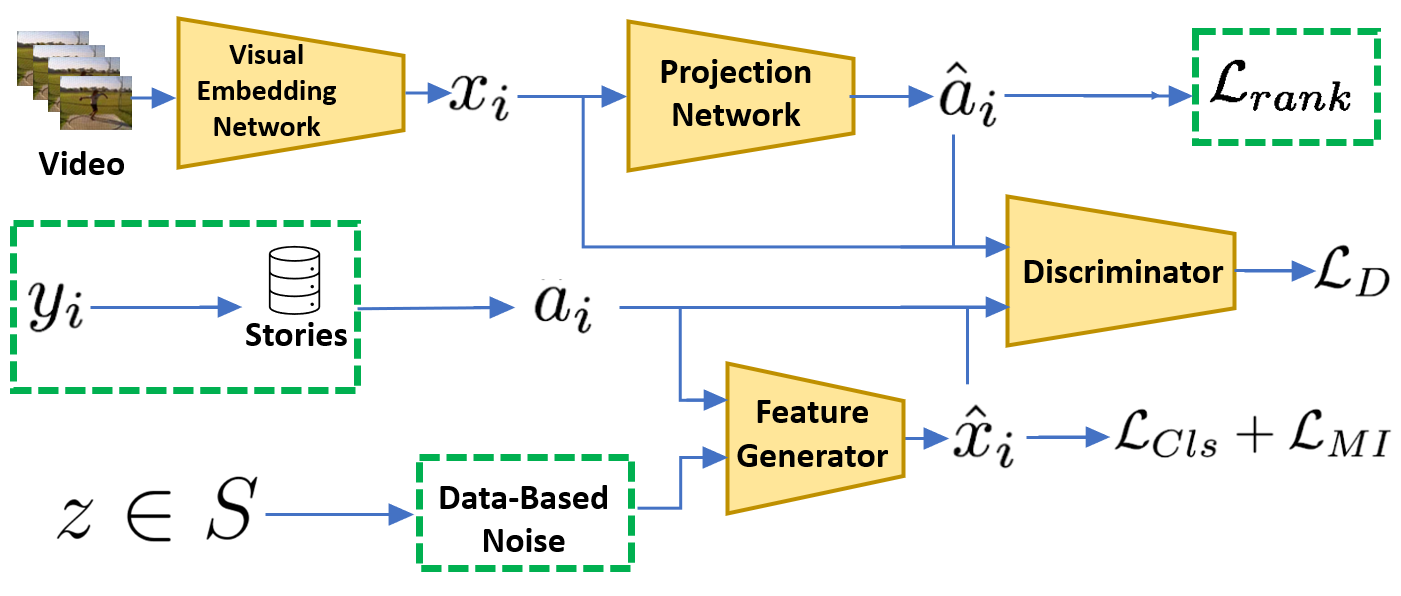}
        \caption{Using Stories~ for feature generation. The elements depicted in yellow are the standard vanilla approach to feature generation for ZS. Depicted in green are the elements that we introduce. 
        }
        \label{fig:generator}
\end{figure*}

\section{Experimental Details}

\label{sec:method}


\subsection{The Zero-Shot and Generalized Zero-Shot Settings}

Let $S$ be the training set of seen classes. $S$ is composed of tuples $\left( x, y, a(y) \right)$, where $x$ represents the spatiotemporal features of a video, $y$ represents the class label in the set of $Y_{S}$ seen class labels, and $a(y)$ denotes the category-specific semantic representation of class $y$, which is either manually annotated or computed automatically, for example using word2vec~\cite{word2vec} or the proposed Stories~. 

Let $U$ be the set of pairs $ \left (u, a(u)\right)$, where $u$ is a class in the set of unseen classes $Y_{U}$ and $a(u)$ are the corresponding semantic representations. The seen $Y_S$ and the unseen classes $Y_U$ do not overlap. 

In the ZS setting, given an input video, the task is to predict a class label in the unseen classes, such as $f_{ZSL}:X\rightarrow Y_{U}$. In the generalized zero-shot (GSZ) setting, given an input video, the task is to predict a class label in the union of the seen and unseen classes, as $f_{GZSL}:X\rightarrow Y_{S}\cup Y_{U}$.

\begin{figure}[!tb]
    \centering
    \begin{minipage}{0.45\textwidth}
        \centering
        \includegraphics[width=\textwidth]{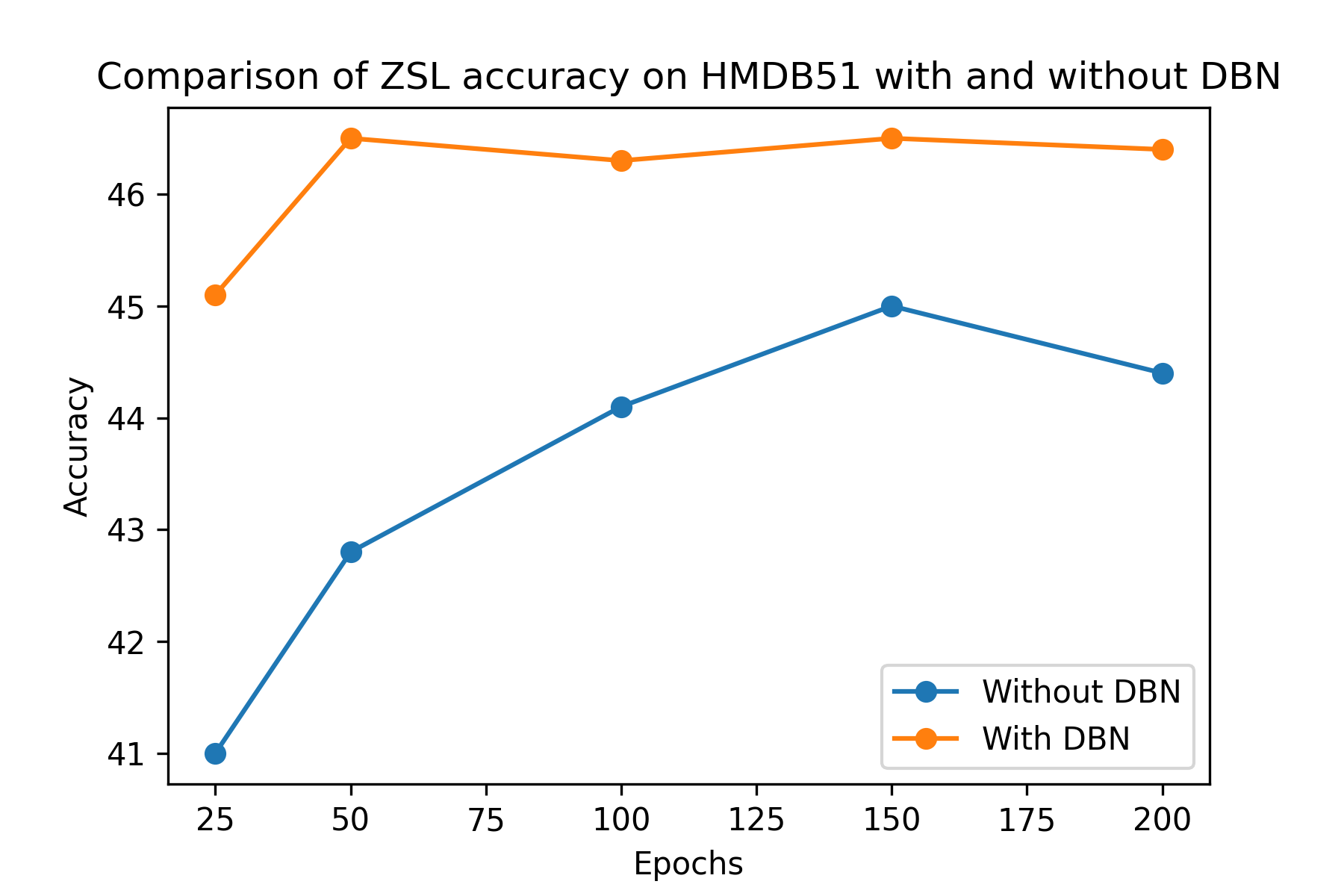}
        \caption{Training the generator using data-driven noise converges much faster than using the standard Gaussian noise.}
        \label{fig:dbn_acc}
    \end{minipage}\hfill
    \begin{minipage}{0.45\textwidth}
        \centering
        \includegraphics[width=\textwidth]{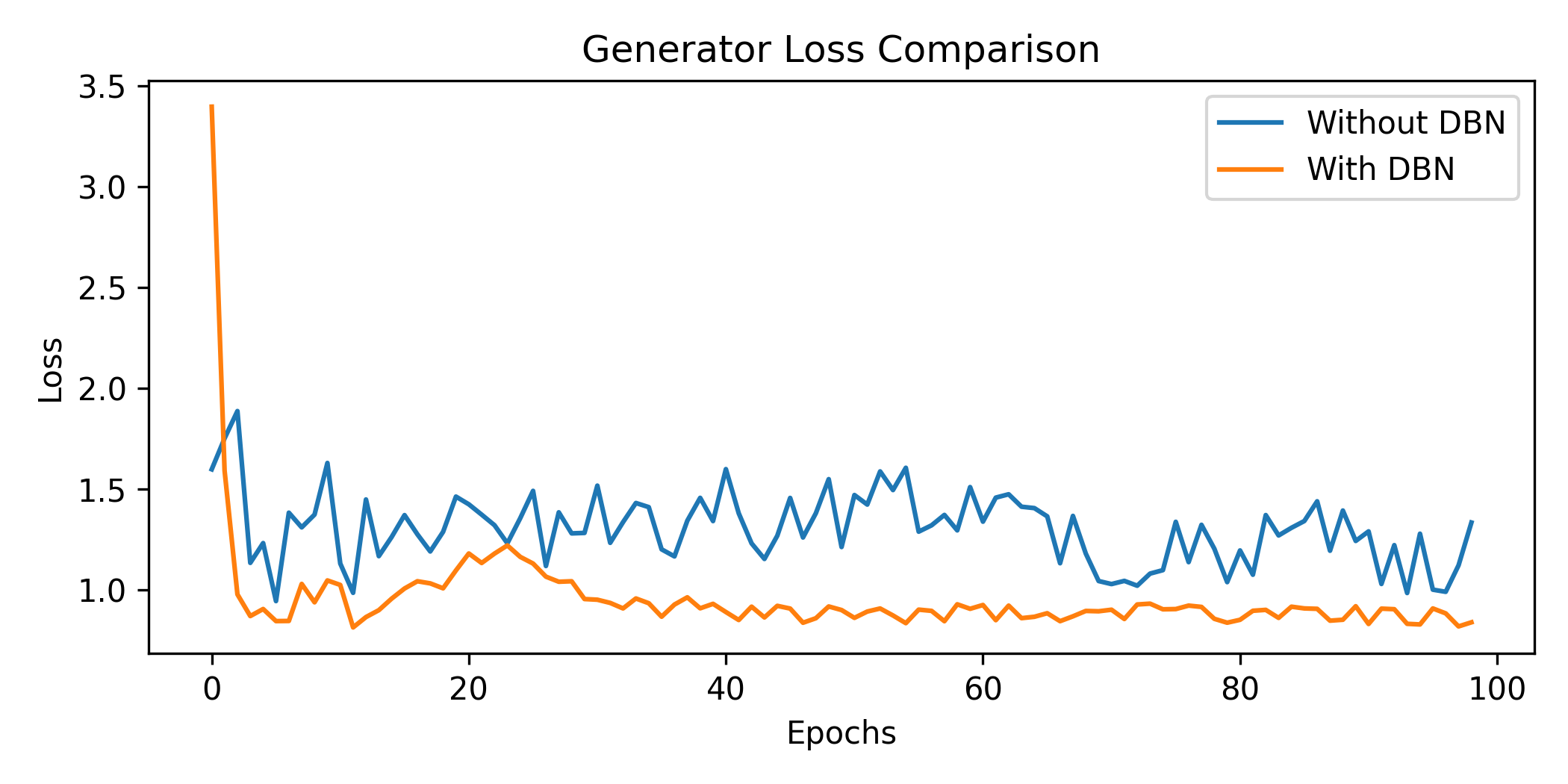}
        \caption{The generator loss using data-driven noise is much more stable, leading to faster convergence and better accuracy.}
        \label{fig:dbn_loss}
    \end{minipage}
\end{figure}

\subsection{Vanilla Feature Generation}


In the standard feature generation pipeline ~\cite{clswgan,OD,GGM2018,vae,han2021inference} the high-level idea is to learn to generate visual features for unseen classes using a GAN~\cite{goodfellow2020generative}, and then, given these synthetic features, train a classifier that takes in visual features and predicts unseen class labels. Figure~\ref{fig:generator} illustrates the overall method.



The GAN comprises a generator ($G$), discriminator ($D$), and projection network ($P$). Generator $G$ creates synthetic visual features ($\hat{x}_i$) from class label semantic embedding ($a_i$) and noise ($z$). $P$ maps visual features ($x_i$) to a semantic embedding approximation ($\hat{a}_i$). $D$ separates real from synthesized features. They're jointly trained via a mini-max game using an optimization function (see Eq~\ref{eq:joint}).


\begin{equation}
\label{eq:joint}
\begin{aligned}
\mathcal{L}_D &= \mathbb{E}_{(x,a)\sim p_{(x_{S}\times a_{S})}}[D(x,P(x))] \\
&\quad - \mathbb{E}_{z\sim p_z}\mathbb{E}_{a\sim p_a}[D(G(a,z),a)] \\
&\quad - \alpha \mathbb{E}_{z\sim p_z}\mathbb{E}_{a\sim p_a}\left[\left(\left\|\nabla_{\hat{x}} D(G(a,z))\right\|_2-1\right)^2\right],
\end{aligned}
\end{equation}

\subsection{Technical Details}
\label{sec:data_noise}

We now describe a few technical choices that we have made to improve the standard pipeline. We present experimental results for some of our choices (including number of nearest neighbors, alternative feature generating models and the need for manual cleaning) and discuss implementation details in the supplementary material.

First, the typical generator takes as input attributes or semantic embeddings and normally distributed noise to generate visual features. The underlying assumption is that the normal distribution can represent all classes. However, this is not necessarily true~\cite{vae}. 

Instead, we use the distribution of the seen classes' features to create the ``noise" for the generator~\cite{vae}, such that the synthetic unseen classes will follow the same distribution. To this end, we use a variational auto-encoder (VAE), which takes as input visual features from the seen classes and it is trained to reconstruct them. 
Once trained, we use the low-dimensional representation of the encoder as the ``noise". 
This simple change in the noise distribution, which we call data-driven noise, benefits in two ways: it improves the overall accuracy and reduces training time by $65\%$ compared to the standard Gaussian noise. See Fig.~\ref{fig:dbn_acc} and Fig.~\ref{fig:dbn_loss}.

Second, one of the risks of learning to generate semantic embeddings (through the ``Projection Network" in Fig.~\ref{fig:generator}) is that synthetic semantic embeddings can be too similar to each other. 
To avoid this, we introduce a ranking loss~\cite{frome2013devise} that pushes apart the generated semantic representation ($\hat{a}_i$) from those of their neighboring classes. Details of this can be found in the supplementary. We refer to this version of the feature generating approach as SDR  (for Stories~, Data-based Noise and Ranking) and we observe that it achieves state-of-the-art results across all datasets and settings.


$p_{(x_{S}\times a_{S})}$ is the joint distribution of visual and semantic descriptors for seen classes, $p_a$ is the empirical distribution of their semantic embeddings, $p_z$ is noise, and $\alpha$ is a penalty coefficient. Additional losses to enhance generated features are the classification regularized loss ($\mathcal{L}_{CLS}$) and the mutual information loss~\cite{belghazi2018mutual} ($\mathcal{L}_{MI}$). These losses form the objective function minimized to train the vanilla pipeline (see Eq~\ref{eq:overall}).
\begin{equation}
\label{eq:overall}
\begin{gathered}
\min_G \min_P \max_D \mathcal{L_D} + \lambda_{1}\mathcal{L}_{CLS}(G) + \lambda_{2}\mathcal{L}_{MI}(G).
\end{gathered}
\end{equation}

Once these networks are trained on the seen classes, the generator is used to synthesize visual features for the unseen classes. The final step is to train a simple classifier using these synthetic visual features as input and the class labels. The loss is the standard cross-entropy loss, and the classifier is a simple multilayer perceptron (MLP).

\subsection{Ablation Study}
\label{sec:ablation}

\setlength{\tabcolsep}{2pt}
\begin{table}[htb]
\begin{center}
\resizebox{0.8\columnwidth}{!}
{
\begin{tabular}{|l|c|c|c|c|c|c|}
\hline 
Stories & DBN & $\mathcal{L}_{rank}$ & $ZSL_{HMDB}$ & $ZSL_{UCF}$ & $GZSL_{HMDB}$ & $GZSL_{UCF}$\\
\hline\hline
$\times$ & $\times$ & $\times$ & 29.1 $\pm$ 3.8 & 37.5 $\pm$ 3.1 & 32.7 $\pm$ 3.4 & 44.4 $\pm$ 3.0  \\
$\times$ & $\times$ & \checkmark & 29.7 $\pm$ 3.5 & 38.0 $\pm$ 3.1 & 35.3 $\pm$ 3.1 & 47.1 $\pm$ 3.2 \\
$\times$ & \checkmark & $\times$ & 30.6 $\pm$ 2.2 & 38.6 $\pm$ 3.4 & 33.3 $\pm$ 3.0 & 44.9 $\pm$ 2.9 \\
\checkmark & $\times$ & $\times$ & 44.6 $\pm$ 2.9 & 60.4 $\pm$ 3.8 & 44.9 $\pm$ 3.6 & 51.0 $\pm$ 2.9 \\
$\times$ & \checkmark & \checkmark & 31.9 $\pm$ 3.2 & 40.9 $\pm$ 2.9 & 35.7 $\pm$ 2.9 & 47.9 $\pm$ 4.1 \\
\checkmark & $\times$ & \checkmark & 45.0 $\pm$ 2.5 & 60.9 $\pm$ 3.5 & 49.0 $\pm$ 3.2 & 54.4 $\pm$ 3.7 \\
\checkmark& \checkmark & $\times$  & 45.9 $\pm$ 2.7 & 61.4 $\pm$ 2.8 & 47.5 $\pm$ 2.6 & 53.7 $\pm$ 3.5 \\
\checkmark & \checkmark & \checkmark & \textbf{46.5 $\pm$ 5.3} & \textbf{61.9 $\pm$ 2.5} & \textbf{49.7 $\pm$ 2.9} & \textbf{54.9 $\pm$ 4.4} \\
\hline
\noalign{\smallskip}
\end{tabular}
}
\end{center}

\caption{Ablation study to explore the impact of each proposed component. }
\label{tbl:ablation}
\end{table}

We propose modifications to the vanilla pipeline of feature generation and based on this, we show the importance of each component here. We see that every proposed contribution benefits over the baseline.  However, crucially, a combination of all three gives us our best results. We note that the improvement from the ranking loss is much more prominent in the GZS setting.

\subsection{Datasets and Evaluation Protocol}

We use the HMDB-51 \cite{hmdb}, UCF-101 \cite{ucf101} and Kinetics \cite{i3d} datasets, as they are the standard choice in ZS action recognition, so that we can compare them with recent state-of-the-art models \cite{gan2016learning,OD,qin2017zero,claster,jigsaw,e2e,x-florence}. 
These datasets contain 783, 6766 and 13320 videos, and have 16, 51 and 101 classes, respectively.

We follow the commonly used 50/50 splits of Xu et al. \cite{xu2017transductive}, where 50\% are the seen classes and 50 are the unseen classes. Similar to previous approaches \cite{zhu2018towards,gan2016learning,qin2017zero,mettes2017spatial,kodirov2015unsupervised}, we report average accuracy and standard deviation over 10 independent runs.  

We also report on the recently introduced TruZe~\cite{truze}. This split accounts for the fact that some classes present on the dataset used for pre-training (Kinetics~\cite{i3d}) overlap with some of the unseen classes in the datasets used in the zero-shot setting. We also report on the Kinetics-220\cite{i3d} split as proposed in ER~\cite{er}. Here, the 220 classes from Kinetics-600~\cite{kinetics600} are treated as unseen classes to a model trained on the Kinetics-400 dataset.

\section{Results}
\label{sec:results}

We now look at the comparison of the proposed method with recent SOTA models in the zero-shot as well as the generalized zero-shot settings. We also report the use of Stories as a general purpose semantic embedding that improves multiple model's performance in a simple plug and play fashion highlighting the benefits of a more extensive semantic embedding in zero-shot action recognition.

\subsection{Zero-Shot Learning Results}
\label{subsec:zsl}

\begin{table*}[htb]
\resizebox{0.99\textwidth}{!}{
{
\begin{tabular}{|c|c|c|c|c|c|c|c|c|c|c|c|c|c|}
\hline
D                            & E   & Bi-Dir GAN              & WGAN                    & OD                      & E2E                     & CLASTER                 & JigSawNet               & ActionCLIP              & ResT                    & X-CLIP                  & \textbf{SDR+I3D}  & \textbf{SDR+CLIP} \\ \hline
\multirow{4}{*}{\textbf{O}} & W   & 40.2 $\pm$ 10.6         & 47.1 $\pm$ 6.4          & 50.5 $\pm$ 6.9          & 61.4 $\pm$ 5.5          & 63.8 $\pm$ 5.7          & 66.4 $\pm$ 6.8          & -                       & -                       & -                       & 55.1 $\pm$ 4.8          & 62.3 $\pm$ 4.5          \\  
                                   & ER  & 54.1 $\pm$ 6.8          & 65.5 $\pm$ 7.2          & 67.6 $\pm$ 6.2          & 66.5 $\pm$ 4.5          & 68.4 $\pm$ 4.1          & 71.5 $\pm$ 6.1          & -                       & -                       & -                       & 69.8 $\pm$ 2.8          & 73.5 $\pm$ 2.9          \\   
                                   & Sto & \textbf{55.5 $\pm$ 6.5} & \textbf{66.2 $\pm$ 7.1} & \textbf{69.1 $\pm$ 5.6} & \textbf{69.9 $\pm$ 5.8} & \textbf{73.1 $\pm$ 6.6} & \textbf{74.9 $\pm$ 5.2} & -                       & -                       & -                       & \textbf{72.5 $\pm$ 2.1} & \textbf{80.1 $\pm$ 2.3} \\   
                                   & SM  & -                       & -                       & -                       & -                       & -                       & -                       & -                       & -                       & -                       & \textbf{74.8 $\pm$ 2.3} & \textbf{82.2 $\pm$ 1.6} \\ \hline
\multirow{4}{*}{H}            & W   & 21.3 $\pm$ 3.2          & 29.1 $\pm$ 3.8          & 30.2 $\pm$ 2.7          & 33.1 $\pm$ 3.4          & 36.6 $\pm$ 4.6          & 35.4 $\pm$ 3.2          & 40.8 $\pm$ 5.4          & 39.3 $\pm$ 3.5          & 43.7 $\pm$ 6.5          & 35.8 $\pm$ 4.7          & 38.9 $\pm$ 3.5          \\ 
                                   & ER  & 25.9 $\pm$ 2.9          & 31.6 $\pm$ 3.1          & 36.1 $\pm$ 2.9          & 36.2 $\pm$ 1.9          & 43.2 $\pm$ 1.9          & 39.3 $\pm$ 3.9          & 44.2 $\pm$ 4.4          & 43.6 $\pm$ 2.9          & 46.6 $\pm$ 6.1          & 41.2 $\pm$ 4.3          & 46.2 $\pm$ 3.1          \\ 
                                   & Sto & \textbf{27.2 $\pm$ 2.7} & \textbf{35.5 $\pm$ 2.8} & \textbf{39.2 $\pm$ 2.8} & \textbf{38.1 $\pm$ 3.6} & \textbf{45.5 $\pm$ 2.6} & \textbf{42.5 $\pm$ 3.2} & \textbf{48.8 $\pm$ 3.2} & \textbf{47.1 $\pm$ 3.5} & \textbf{50.1 $\pm$ 6.1} & \textbf{46.8 $\pm$ 5.0} & \textbf{52.7 $\pm$ 3.4} \\ 
                                   & SM  & -                       & -                       & -                       & -                       & -                       & -                       & -                       & -                       & -                       & \textbf{48.9 $\pm$ 4.4} & \textbf{54.4 $\pm$ 4.1} \\ \hline  
\multirow{4}{*}{U}            & W   & 21.8 $\pm$ 3.6          & 25.8 $\pm$ 3.2          & 26.9 $\pm$ 2.8          & 46.2 $\pm$ 3.8          & 46.7 $\pm$ 5.4          & 53.3 $\pm$ 3.1          & 58.3$\pm$ 3.4           & 58.7 $\pm$ 3.3          & 70.1 $\pm$ 3.4          & 41.5 $\pm$ 2.5          & 44.8 $\pm$ 4.2          \\ 
                                   & ER  & 28.0 $\pm$ 3.4          & 37.9 $\pm$ 2.5          & 42.4 $\pm$ 3.4          & 52.4 $\pm$ 3.3          & 53.9 $\pm$ 2.5          & 56.8 $\pm$ 2.8          & 64.3 $\pm$ 3.8          & 62.6 $\pm$ 4.1          & 72.2 $\pm$ 2.3          & 50.3 $\pm$ 1.1          & 54.2 $\pm$ 3.5          \\ 
                                   & Sto & \textbf{29.5 $\pm$ 3.2} & \textbf{40.1 $\pm$ 3.7} & \textbf{50.3 $\pm$ 3.0} & \textbf{55.1 $\pm$ 3.3} & \textbf{59.6 $\pm$ 2.8} & \textbf{61.6 $\pm$ 3.5} & \textbf{71.8 $\pm$ 2.7} & \textbf{65.3 $\pm$ 2.5} & \textbf{73.8 $\pm$ 2.9} & \textbf{62.9 $\pm$ 1.6} & \textbf{73.4 $\pm$ 2.7} \\ 
                                   & SM  & -                       & -                       & -                       & -                       & -                       & -                       & -                       & -                       & -                       & \textbf{64.9 $\pm$ 2.1} & \textbf{75.5 $\pm$ 3.2} \\ \hline
\end{tabular}

}}
\vspace{0.3cm}
\caption{Results on ZSL. E: semantic embedding, W: word2vec embedding, ER: Elaborate Rehearsals, Sto: Stories. SM corresponds to the single model training. We use the datasets Olympics (O), HMDB51  (H) and UCF101 (U).}
\label{tab:zsl}
\end{table*}

We look at the use of \textit{Stories} as semantic embedding to a wide variety of models here, from older models all the way to the most recent ones in the ZS literature. We use I3D  and CLIP-based features~\cite{x-florence} to compare their effect. We list these results as SDR+I3D and SDR+CLIP respectively. We observe an improvement across all of them and across all datasets, demonstrating that \textit{Stories} is clearly model agnostic. Results can be seen in Table~\ref{tab:zsl}.


We also observe that the proposed changes to the vanilla feature generation method which we call SDR  consistently outperforms all approaches across all datasets, achieving a new state-of-the-art. We experiment with using a single model for all datasets, by training on Kinetics and not doing any fine-tuning for the smaller datasets. This is the last row of the table, which we call ``SDR  (Ours) + SM". It is remarkable and quite promising that, without the need to fine-tune, this single model achieves even better performance. 

We also evaluate on the Kinetics-220 dataset as proposed in ER~\cite{er}. There are fewer methods who report on this split, but it is interesting as it is much larger. Results can be seen in Table~\ref{tbl:results:zsl_k200}. We observe that the proposed SDR  outperforms all previous work. We see significant gains of up to 5\%. 

\begin{table}
\centering
\begin{tabular}{|l|c|c|}
\hline
Method & Top-1 Acc & Top-5 Acc\\
\hline\hline
DEVISE~\cite{frome2013devise} & 23.8 $\pm$ 0.3 & 51.0 $\pm$ 0.6\\
SJE~\cite{SJE} & 22.3 $\pm$ 0.6 &  48.2 $\pm$ 0.4 \\
ER~\cite{er} & 42.1 $\pm$ 1.4 & 73.1 $\pm$ 0.3 \\
JigSawNet~\cite{jigsaw} & 45.9 $\pm$ 1.6 & 78.8 $\pm$ 1.0 \\
\textbf{SDR+I3D} & \textbf{50.8 $\pm$ 1.9} & \textbf{82.9 $\pm$ 1.3} \\
\textbf{SDR+CLIP} & \textbf{55.1 $\pm$ 2.2} & \textbf{86.1 $\pm$ 3.1} \\
\hline
\end{tabular}%
\caption{Results of ZS in Kinetics-220.}
\label{tbl:results:zsl_k200}
\end{table}

Finally, we evaluate on the stricter TruZe~\cite{truze} split that ensures no overlap between pre-trained model and test classes. Results are shown in Table~\ref{tbl:truze}. We report the mean class accuracy in ZS setting and harmonic mean of seen and unseen class accuracies in GZS. The split refers to the train/test split used.

\begin{table}
\centering
\begin{tabular}{| *{7}{c|} }
\hline
Method & \multicolumn{3}{c|}{UCF101 } & \multicolumn{3}{c|}{HMDB51}\\
 & Split & ZSL & GZSL & Split & ZSL & GZSL \\
 \hline\hline
WGAN & 67/34 & 22.5 & 36.3 & 29/22 & 21.1 & 31.8 \\
OD & 67/34 & 22.9 & 42.4 & 29/22 & 21.7 & 35.5 \\
CLASTER & 67/34 & 45.8 & 47.3 & 29/22 & 33.2 & 44.5 \\
\textbf{SDR+I3D} & 67/34 & \textbf{49.7} & \textbf{51.3} & 29/22 & \textbf{34.9} & \textbf{45.5} \\
\textbf{SDR+CLIP} & 67/34 & \textbf{53.9} & \textbf{56.2} & 29/22 & \textbf{38.7} & \textbf{49.5} \\
\hline
VCAP~\cite{tellme} & 0/34 & 49.1 & - & 0/22 & 20.4 & - \\
\textbf{SDR+I3D SM} & 0/34 & \textbf{51.5} & \textbf{52.2} & 0/22 & \textbf{36.1} & \textbf{46.6} \\
\textbf{SDR+CLIP SM} & 0/34 & \textbf{55.1} & \textbf{57.7} & 0/22 & \textbf{40.8} & \textbf{51.1} \\
\hline
\end{tabular}
\caption{Results of ZS and GZS on TruZe. }
\label{tbl:truze}
\end{table}


\begin{table*}[htb]
\begin{center}

\resizebox{0.99\textwidth}{!}{
{
\begin{tabular}{|c|c|c|c|c|c|c|c|c|}
\hline
Dataset                   & E   & Bi-Dir GAN              & GGM                      & WGAN                    & OD                      & CLASTER                 & \textbf{SDR+I3D}        & \textbf{SDR+CLIP}       \\ \hline
\multirow{4}{*}{Olympics} & W   & 44.2 $\pm$ 11.2         & 52.4 $\pm$ 12.2          & 59.9 $\pm$ 5.3          & 66.2 $\pm$ 6.3          & 69.1 $\pm$ 5.4          & 67.1 $\pm$ 4.2          & 69.7 $\pm$ 4.1          \\  
                          & ER  & 53.6 $\pm$ 6.2          & 59.1 $\pm$ 12.1          & 63.7 $\pm$ 6.6          & 69.7 $\pm$ 6.5          & 72.5 $\pm$ 3.5          & 70.4 $\pm$ 2.3          & 75.5 $\pm$ 3.3          \\  
                          & Sto & \textbf{55.9 $\pm$ 4.2} & \textbf{59.9 $\pm$ 11.6} & \textbf{67.1 $\pm$ 5.1} & \textbf{72.4 $\pm$ 4.9} & \textbf{74.9 $\pm$ 6.1} & \textbf{74.5 $\pm$ 3.9} & \textbf{79.5 $\pm$ 3.5} \\  
                          & SM  & -                       & -                        & -                       & -                       & -                       & \textbf{76.6 $\pm$ 3.6} & \textbf{81.1 $\pm$ 3.0} \\ \hline
\multirow{4}{*}{HMDB51}   & W   & 17.5 $\pm$ 2.4          & 20.1 $\pm$ 2.1           & 32.7 $\pm$ 3.4          & 36.1 $\pm$ 2.2          & 48.0 $\pm$ 2.4          & 36.3 $\pm$ 5.1          & 39.5 $\pm$ 3.1          \\  
                          & ER  & 26.1 $\pm$ 2.3          & 28.2 $\pm$ 3.3           & 35.2 $\pm$ 3.5          & 38.1 $\pm$ 2.4          & 50.8 $\pm$ 2.8          & 42.1 $\pm$ 4.5          & 45.4 $\pm$ 3.6          \\  
                          & Sto & \textbf{27.7 $\pm$ 2.1} & \textbf{29.1 $\pm$ 2.9}  & \textbf{38.5 $\pm$ 3.3} & \textbf{40.9 $\pm$ 3.8} & \textbf{53.5 $\pm$ 2.4} & \textbf{49.7 $\pm$ 2.9} & \textbf{53.5 $\pm$ 3.3} \\  
                          & SM  & -                       & -                        & -                       & -                       & -                       & \textbf{50.9 $\pm$ 2.6} & \textbf{56.1 $\pm$ 3.2} \\ \hline
\multirow{4}{*}{UCF101}   & W   & 22.7 $\pm$ 2.5          & 23.7 $\pm$ 1.2           & 44.4 $\pm$ 3.0          & 49.4 $\pm$ 2.4          & 51.3 $\pm$ 3.5          & 41.9 $\pm$ 2.7          & 44.3 $\pm$ 4.6          \\  
                          & ER  & 26.2 $\pm$ 4.2          & 26.5 $\pm$ 2.5           & 46.1 $\pm$ 3.5          & 53.2 $\pm$ 3.1          & 52.8 $\pm$ 2.1          & 45.5 $\pm$ 1.1          & 49.1 $\pm$ 2.9          \\  
                          & Sto & \textbf{29.1 $\pm$ 3.4} & \textbf{27.7 $\pm$ 3.1}  & \textbf{48.3 $\pm$ 3.2} & \textbf{55.5 $\pm$ 3.3} & \textbf{54.1 $\pm$ 3.3} & \textbf{54.9 $\pm$ 4.4} & \textbf{57.8 $\pm$ 4.1} \\  
                          & SM  & -                       & -                        & -                       & -                       & -                       & \textbf{57.2 $\pm$ 3.5} & \textbf{59.7 $\pm$ 3.1} \\ \hline
\end{tabular}

}}
\caption{Results on GZSL. E: semantic embedding, W: word2vec embedding, ER: Elaborate Rehearsals, Sto: Stories. SM corresponds to the single model training.}

\label{tab:gzsl}
\end{center}
\end{table*}

\subsection{Generalized Zero-Shot Learning Results}

Generalized ZS action recognition is less explored in comparison to the ZS setting. Nonetheless, we choose models such as Bi-Dir GAN, GGM, WGAN, OD and CLASTER as recent state-of-the-art methods evaluating on this setting. We train an out-of-distribution detector (OOD) following OD~\cite{OD} and two separate classifiers for the seen and unseen classes along with the OOD network.

Table~\ref{tab:gzsl} shows the results, with the harmonic mean of the seen and unseen class accuracies. Similar to the zero-shot case, we use both I3D and CLIP-based~\cite{x-florence} backbones and list these results as SDR+I3D and SDR+CLIP respectively. We follow the earlier approach of using different semantic embeddings to show the performance gain that using SDR  gives us. We provide a detailed seen and unseen class accuracy comparison in the supplementary material.

\section{Discussion}

\subsection{Why Does Using a Single Model Work?}
\label{sec:single}

One curious question to ponder would be why the single model trained on a large dataset like Kinetics-400~\cite{i3d} results in better performance than the models fine-tuned on the smaller datasets. Our hypothesis is that the feature generator trained on a larger dataset has a better distribution of data to learn from as the data-driven noise that we use is more representative of the real visual world. 
This in turn generates more realistically distributed features, which in turn results in the improved performance.

\subsection{Possible Limitations}

One possible limitation of our approach is that the Stories~ may focus on one specific way of performing each action, while other valid methods may exist to do the same action. For example, the story for "shuffling cards" details the riffle shuffling technique, but other shuffling techniques scuh as the overhand shuffling could occur in videos of this class. Similarly, some parts of the stories may describe non-visual aspects like memorizing lines for the "acting in play" class that are not depicted in the videos. Our current approach does not explicitly address potential mismatches between the textual stories and visual contents. 

Still, we believe non-visual cues actually help making the semantic embeddings more distinct, as these cues are unique to each class's story. However, this remains a limitation worth noting. One area of future work is exploring how to make the stories more comprehensive by incorporating multiple variations of actions. We also plan to investigate techniques for identifying and excluding non-visual sentences that do not translate to visual features. Overall, handling the diversity of real videos compared to procedural descriptions remains an open challenge that we aim to address.

\section{Conclusion}

The introduction of the novel Stories dataset provides rich textual narratives that establish connections between diverse action classes. Leveraging this contextual data enables modeling of nuanced relationships between actions, overcoming previous limitations in ZS action recognition. Stories enables significant improvement for multiple SOTA models. Building on Stories, our proposed feature generating approach harnessing Stories achieves new state-of-the-art performance on multiple benchmarks without any target dataset fine-tuning. This demonstrates the value of Stories as a resource to enable ZS transfer and significant progress in video understanding without reliance on large labeled datasets.

%
%
\bibliographystyle{splncs04}
\bibliography{main}
\end{document}